\documentclass[a4paper,conference]{IEEEtran}
\ifCLASSINFOpdf
  \usepackage[pdftex]{graphicx}
  \graphicspath{{./charts/}}
  \usepackage{subfigure}
  \usepackage{amsmath}
  
\else
\fi
\usepackage{caption2}
\usepackage{multirow}
\usepackage{cite}
\usepackage{CJK}
\usepackage{array}
\usepackage{diagbox}
\begin{document}
%


\title{Quality Classified Image Analysis with Application to Face Detection and Recognition}

\author{\IEEEauthorblockN{Fei Yang}
\IEEEauthorblockA{International Doctoral Innovation Centre\\
	University of Nottingham Ningbo China\\
}
\and
\IEEEauthorblockN{Qian Zhang}
\IEEEauthorblockA{School of Computer Science\\
	University of Nottingham Ningbo China\\
}

\and
\and
\IEEEauthorblockN{Miaohui Wang and Guoping Qiu}
\IEEEauthorblockA{College of Information Engineering\\
Shenzhen University,
Shenzhen, China \\
}}


%


\maketitle

\begin{abstract}

Motion blur, out of focus, insufficient spatial resolution, lossy compression and many other factors can all cause an image to have poor quality. However, image quality is a largely ignored issue in traditional pattern recognition literature. In this paper, we use face detection and recognition as case studies to show that image quality is an essential factor which will affect the performances of traditional algorithms. We demonstrated that it is not the image quality itself that is the most important, but rather the quality of the images in the training set should have similar quality as those in the testing set. To handle real-world application scenarios where images with different kinds and severities of degradation can be presented to the system, we have developed a quality classified image analysis framework to deal with images of mixed qualities adaptively. We use deep neural networks first to classify images based on their quality classes and then design a separate face detector and recognizer for images in each quality class. We will present experimental results to show that our quality classified framework can accurately classify images based on the type and severity of image degradations and can significantly boost the performances of state-of-the-art face detector and recognizer in dealing with image datasets containing mixed quality images.

\end{abstract}


%
\IEEEpeerreviewmaketitle

\section{Introduction}

\begin{figure}{}
	\centering
	\includegraphics[angle=0,scale = 0.4]{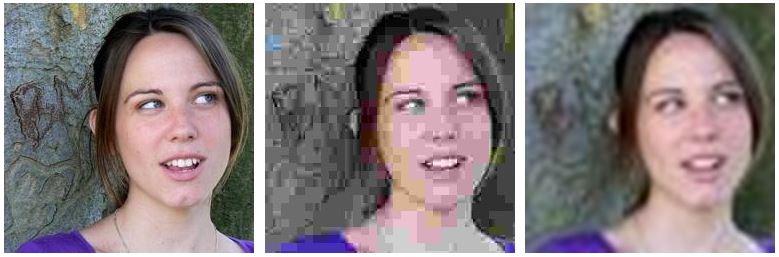}
	\centering
	\caption{A good quality image (left), its low quality JPEG compressed version (middle) and its low spatial resolution version (right)}.
	\label{decreasesamples}
\end{figure}

Object detection and recognition have achieved significant progress in recent years. In real-world application scenarios, motion blur, lossy image compression, insufficient spatial resolution caused by out of focus or objects being too far away and other factors can all result in poor image quality problems. Figure \ref{decreasesamples} shows two typical low-quality versions of an image caused by lossy image compression and low spatial resolution. Surprising, in the literature, image quality issue is mostly ignored, and authors almost always implicitly assume that the quality of images they are dealing with is good and not an issue. In this paper, we will explicitly show that image quality is an important factor that will affect the performances of object recognition and detection algorithms.  As well as recognizing the image quality issue in image analysis, we also propose an innovative solution for explicitly taking into account the image quality in designing object detection and recognition systems.

\begin{figure}{}
	\centering
	\includegraphics[height=4.3cm,width=8cm,angle=0,scale=1]{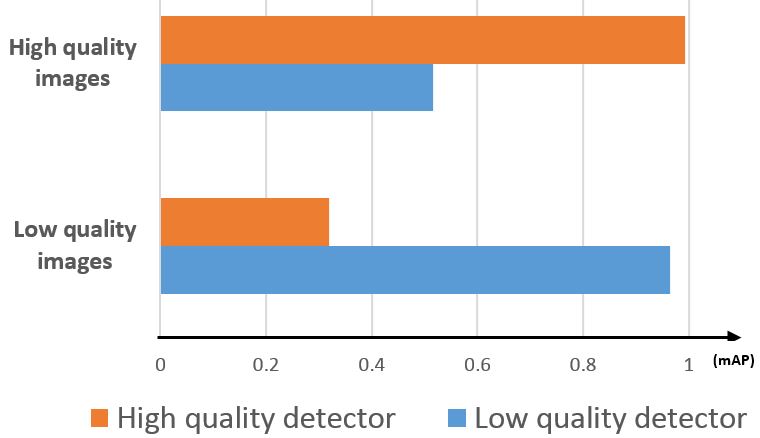}
	\centering
	\caption{Face detection performances (mAP - mean average precision) for detectors trained and tested on images of different qualities.}
	\label{highlowmodeltest}
\end{figure}

We begin by presenting a motivating experiment. We first take a publicly available face detection dataset \cite{koestinger2011annotated}, and then downscale each image in the dataset from $512\times 512$ into $40\times 40$ pixels, resulting in a low-quality dataset. We then take one of the latest deep learning based face detection techniques \cite{Sun2017Face} and train a high image quality detector (using the original resolution images) and a low image quality detector (using the 40 by 40 pixels images). We then test the two detectors on both high and low-quality images. The results are shown in Figure \ref{highlowmodeltest}. It is seen that the high image quality detector works very well on the high-quality testing images; however, its performance is much poorer for the low-quality testing images. Similarly, it can be seen that the low image quality detector performs very well on the low-quality testing images, but its performance deteriorates significantly for the high-quality images. This example tells us, it is not the image quality itself that is the most important in designing a good face detector, but rather the quality of the images used in training the detector should be similar to that of images in the testing set. In many ways, this is to be expected and also makes good sense, nevertheless, this motivating experiment has confirmed that image quality is an important issue needs to be considered in designing an image analysis solution.   
 
Using face detection and recognition as specific applications, we have developed a novel quality classified image analysis framework. To give this study a better focus, we consider two specific types of image quality issues, one caused by image compression (specifically JPEG compression), and the other caused by low spatial resolution. 

Partly motivated by the results in Figure \ref{highlowmodeltest}, which suggests that it is not the image quality itself that is the most important, but rather the quality of the images used to train the analyzer should be similar to that of those on which the analyzer will be tested. Based on this observation, our strategy is first to classify the input images into different quality classes and then designs a suitable image analyzer for an individual image quality class.  

We make following contributions in this paper. First, we have developed an image quality classified face detection and recognition framework which can better handle images of mixed qualities. Second, we have developed a deep learning neural network based image quality classifier and show that we can first determine if an image is of good or poor quality; and then for poor quality images, we can not only determine whether it is caused by JPEG compression or by low spatial resolution, but also the severity of compression and down-sampling. Third, we have developed a method that first designs separate object detectors or classifiers for different classes of image quality in the training stage, and then in the testing stage, automatically sends an image to the first few most suitable individual detectors or classifiers whose outputs are then fused together to improve performances. We demonstrate that our new image quality classified face detection and recognition approach improves state-of-the-art methods for handling mixed image quality datasets.  


\section{Related Works}

The image quality estimation problem has been studied for a long time in the area of image processing. It learns the visual difference caused by image quality, such as lossy compression, brightness, sharpness, and resolution. Several convolutional neural networks based methods have been developed to assess the quality of whole image \cite{Kang2015Simultaneous,Fu2016Blind}. In the research of Image Quality Assessment (IQA)\cite{Marchesotti2012Image}, a rating score is obtained by solving a regression problem. However, all the quality scores are labeled by human beings \cite{Hou2015Blind,Horita2004No}, which is subjective. And all the images are labeled discarding their specific quality classes, resulting in the images with the same quality score containing different quality classes and visual appearance, which also makes the network hard to converge. Meanwhile, researchers are also conducted in the field of JPEG compression related assessment. \cite{Pevny2008Detection} focused on detecting whether one image is compressed with JPEG. Further research including estimating the quantization table \cite{lukavs2003estimation}, 
and removing blocking artifacts \cite{Fan2003Identification}.
These methods either rely on external information from header file or special designed hand-crafted features for detecting blocking artifacts. To our knowledge, no one has ever used CNN based methods to estimate the detailed quality information.

There are only a few works that concern the effects of image quality in solving detection or recognition problems \cite{dodge2016understanding, Karahan2016How,Liu2017Quality,Haque2015Quality}. Two types of research work are summarized. One is to analyze how much the image quality can affect the performance of standard object or face problems\cite{dodge2016understanding, Karahan2016How}. They compare the model robustness by testing the models on manually decreased low-quality images. 

The other is to develop methods to overcome the low-quality problem in real application scenarios through identifying the low-quality images discarding their quality classes and the corresponding severity. \cite{Liu2017Quality} proposed a quality assessment network within an end-to-end training framework in human re-id and face recognition problems. Instead of a single image, the network regards a set of images or a sequence of images as a recognition subject entity and handles the set to set recognition by predicting the quality score of the image within each set. A low-quality image is given a small score and, hence, reducing its impact on the whole set. Similarly, \cite{Haque2015Quality} concerned the image quality problem in facial landmark detection by selecting the high-quality image in a video sequence. The low-quality frame problem is addressed by locating and replacing with high-quality face in the previous video frames. They also narrowed the quality causes within face patches, which assumes the face can be correctly detected under poor image quality.
 
Our proposed method belongs to the second type. However, instead of purely identifying and discarding the low-quality images, our main strength is to predict the quality classes as well as their severity explicitly and handle them differently with the specific prior knowledge. It can be widely adopted to handle unknown quality images any image-based detection and recognition problems, without an additional requirement for manually labeled data.

\begin{figure*}{}
	\centering
	\includegraphics[angle=0,scale=0.28]{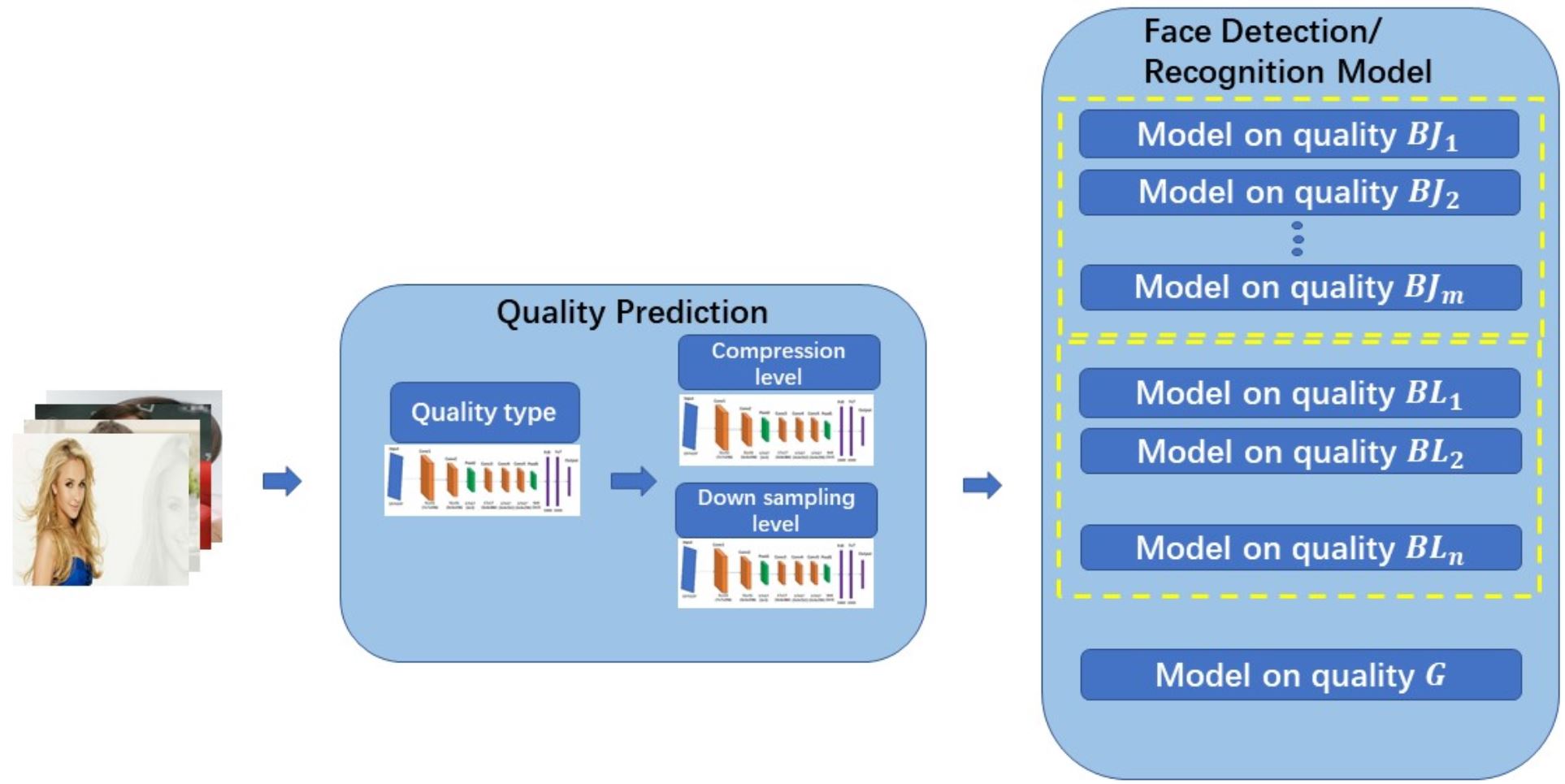}
	\centering
	\caption{Overall framework for quality level prediction and quality classified image face detection/recognition.}
	\label{overallframeworkface}
\end{figure*}



\section{Quality Classified Image Analysis}
\subsection{Overall framework}
We are interested in the problem of dealing with input images of unknown quality and use face detection/recognition as specific case studies. In particular, we consider two types of image quality problems, JPEG compression, and low-resolution. For each type of quality issue, we also consider the severity of the quality issue, and we call this the quality level. We first define following three main image quality classes: Good Quality ($G$), Bad Quality JPEG compression ($BJ$) and Bad Quality low resolution ($BL$). For the $BJ$ and $BL$ class, we define two subsets based on the level of severity of compression or low-resolution, $\{BJ_i, i = 1, 2, 3, ..., m\}$ and $\{BL_j, j = 1, 2, 3, ...,n\}$. Therefore, an image can be classified based on its quality into one of the classes in the following quality class set $C=\{G, BJ_i, BL_j; i =1, 2, ...m, j = 1, 2, ...n\}$. 

As demonstrated in the motivating experiment (see Figure \ref{highlowmodeltest}), image quality is an essential issue in image analysis. The question is how should we deal with it. 

We take a quality classification approach and classify an image into one of the quality classes we just defined. Once we have determined the quality class of an image, we can then design an image analyzer suitable for that image quality class. Our solution framework is shown in Figure \ref{overallframeworkface}. We first use a deep learning network to class an image into one of the three first-level quality classes $\{G, BJ,BL\}$, then, we use another deep learning network to classify those in the $BJ$ class into their subclasses $\{BJ_i, i = 1, 2, 3, ..., m\}$, and a third deep learning network to classify those in the $BL$ class into their subclasses $\{BL_j, j = 1, 2, 3, ..., n\}$. 

We train multiple detection/recognition models using images of different quality levels. For a given input image, we select the first few most likely quality classes the input belongs and fuse the outputs of the models corresponding to these classes. 


\subsection{Quality prediction network}
\label{sec:qualitypredictionnetwork}
\begin{figure}{}
	\centering
	\includegraphics[angle=0,scale=0.3]{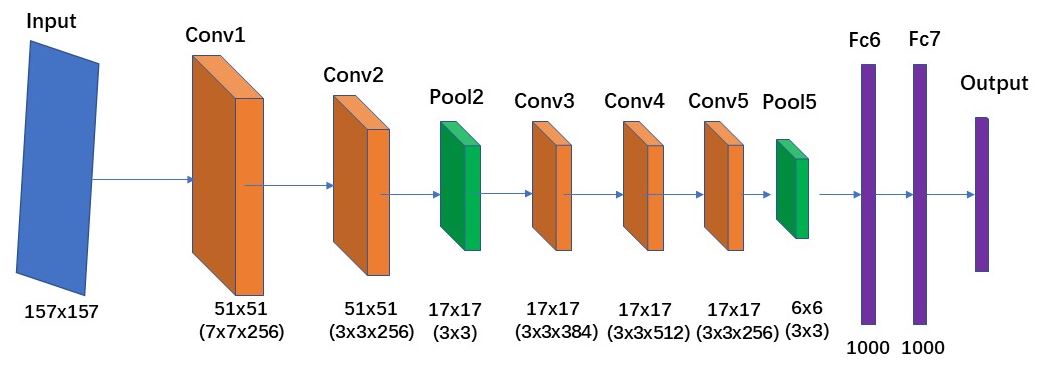}
	\centering
	\caption{Quality prediction network architecture.}
	\label{predictionnet}
\end{figure}

All the three prediction networks share similar network architecture. Figure. \ref{predictionnet} illustrates the details of our proposed quality prediction convolutional neural network. We randomly crop image patches from the input image. The image patch size is set as $157\times157$. Similar to a typical CNN, we stack five convolutional and two fully connected layers. To reduce the feature dimension, we put a 3-stride pooling layer after the second and fifth convolutional layers, respectively. Two fully connected layers (Fc6 and Fc7), containing 1000 neurons each, are followed by the final pooling layer. A Softmax output is used to generate the first-level class scores $\{p(G),p(BJ),p(BL)\}$, as well as the second-level class scores, $\{p(BJ_i),i=1,2,3,...,m\}$ and $\{p(BL_j),j=1,2,3,...,n\}$. 
\begin{figure*}
	\centering
	\includegraphics[angle=0,scale=0.30]{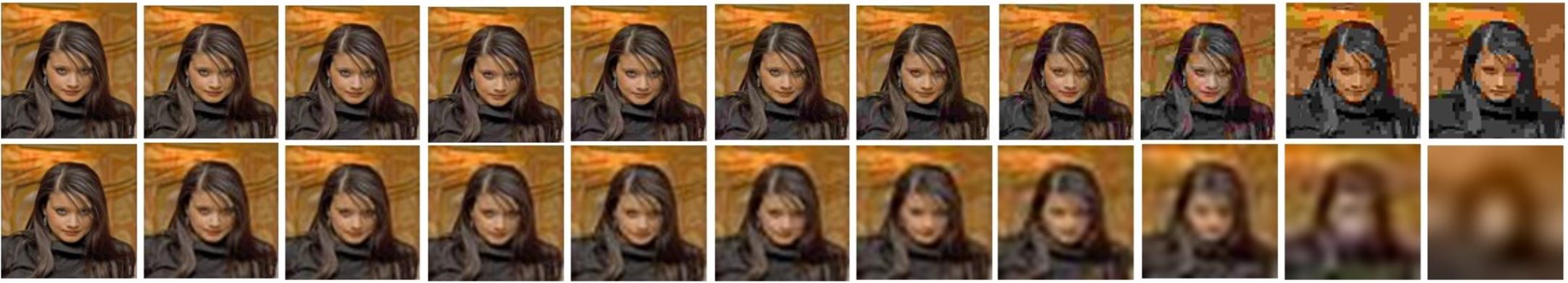}
	\caption{The first row contains the JPEG compression level samples in the setting of $\{uncompressed,27,24,21,18,15,12,9,6,3,0\}$. The second row contains down-sampling level samples in the setting of $\{unresized, 80*80, 72*72, 64*64, 56*56,48*48, 40*40, 32*32, 24*24, 16*16,8*8 \}$.}
	\label{classsamples}
\end{figure*}

\subsection{Target model selection and result fusion}\label{sec:fusion}

The class score vectors $p(\cdot)$, which come from the output of the three quality prediction networks, is fused to generate a single quality score vector $P_{c}$ with 

\begin{equation*}
\begin{split}
P_{C}=\{p(G)*1, p(BJ)*p(BJ_i),p(BL)*p(BL_j),\\i = 1, 2, 3, ..., m,j = 1, 2, 3, ..., n\}
\end{split}
\end{equation*}
$P_C$ indicates the probability of the input image patch belonging to each quality class, according to which, we select the top $K$ corresponding trained models to form the final image analyzer.

In a quality classified face detection application, we merge the series of face bounding boxes produced by the top $K$ face detection models, where the models are trained on different quality-level datasets using an existing face detection method. A Non Maximum Suppression \cite{Neubeck2006Efficient} method is applied to locate the redundant boxes. 

Similarly, in a quality classified face recognition application, we aggregate face identity scores coming from the top $K$ face recognition models with their weight calculated according to $P_C$.

\section{Experiments}
We conduct two sets of experiments. The first one present the quality prediction results to show how well our proposed network can learn quality feature and accurately predict the quality classes. In the second one, we evaluate our proposed quality classified image analysis framework to demonstrate its effectiveness 
in face detection and recognition applications.

\subsection{Quality Prediction}
\subsubsection{\textbf{Dataset}} We randomly selected 10,000 images from COCO \cite{lin2014microsoft} and MegaFace \cite{nech2017level} separately and processed them into different quality classes with JPEG compression or down-sampling. The quality prediction network is trained on COCO images and finetuned on MegaFace images. We compressed each image by JPEG standard with 11 quality factors = $\{27,24,21,18,15,12,9,6,3,0\}$ and down sampled each image into 11 classes with sizes = $\{80*80, 72*72, 64*64, 56*56,48*48, 40*40, 32*32, 24*24, 16*16,8*8 \}$, respectively.

\subsubsection{\textbf{Quality type prediction}}
The first quality prediction network is responsible for predicting the low-quality types, which is trained on a set with three classes, good images $G$, JPEG compressed low-quality images $BJ$ and down-sampling low-quality images $BL$, denoted as $\{G,BJ,BL\}$. 
We split the dataset into training(80\%) and testing(20\%) sets, the testing result reached a very high accuracy of 99.9\%. Note that instead of image's quality type, we place more emphasis on the resulting probability, which indicates the relative weightings of each quality type contributing to the final results during fusion.
 
\subsubsection{\textbf{Quality level prediction}}
We predict the exact quality severity according to the predefined quality class levels, JPEG compressed levels and down-sampling levels. As a comparison, we also tested other popular CNN architectures, including AlexNet, Inception, VGG, and ResNet. We show the overall results in TABLE.\ref{qualityprediction}. 

From TABLE.\ref{qualityprediction}, we can see our proposed network (with five Conv. layers) can obtain a reasonably high accuracy while costing less computation time. We further evaluated the proposed network with only two convolutional layers and found that the performance dropped while the feed-forward time increased. The reason is that Conv2 layer generates a larger feature map, resulting in more computations between the feature map and fully connected layer. We also show the detailed results in the confusion matrix in TABLE \ref{confmatrixquality}. It is seen that all the predictions are classified into the correct classes or the quality level very close to the correct classes, which shows the proposed CNN model has learned the quality features well and would work well in our framework.

\begin{table}
	\centering
	\footnotesize
	\newcommand{\tabincell}[2]{\begin{tabular}{@{}#1@{}}#2\end{tabular}}
	\begin{tabular}{|c|c|c|c|} 
		\hline  
		Network       & \tabincell{c}{JPEG \\ compression\\level \\prediction} & \tabincell{c}{Down \\sampling\\ level \\ prediction} & \tabincell{c}{Feed-\\forward \\time} \\ \hline
		AlexNet       & 77.2\%             & 84.6\%   & 58.4ms   \\ \hline
		Inception V3  & 87.3\%             & 94.2\%   & 95.9ms    \\ \hline
		VGG-16        & 90.6\%             & 96.8\%   & 112.9ms   \\ \hline
		ResNet-50     & 91.5\%             & 98.2\%   & 72.7ms  \\ \hline
		ResNet-101    & Not Converge       & Not Converge & 177.8ms \\ \hline
		\tabincell{c}{Proposed net \\ (Two Conv. layers)}   & 76.4\%  & 84.3\%  &58.0ms    \\ \hline
		\tabincell{c}{Proposed net \\ (Five Conv. layers)}  & 87.8\%  & 95.2\%   & 18.5ms     \\ \hline
	\end{tabular}
	\caption{	\footnotesize Quality level prediction accuracy and model testing time}\label{qualityprediction}
\end{table}


\begin{table*}
\scriptsize
\centering
	\subtable[JPEG compression level class] {
		\begin{tabular}{|p{0.3cm}<{\centering}|p{0.3cm}<{\centering}|p{0.3cm}<{\centering}|p{0.3cm}<{\centering}|p{0.3cm}<{\centering}|p{0.3cm}<{\centering}|p{0.3cm}<{\centering}|p{0.3cm}<{\centering}|p{0.3cm}<{\centering}|p{0.3cm}<{\centering}|p{0.3cm}<{\centering}|p{0.3cm}<{\centering}|}
			\hline
			\diagbox{}{}  & 1   & 2   &3   & 4  & 5  & 6  & 7  & 8  & 9  & 10  & 11 \\ \hline
			1 & 2000 & 0  & 0  & 0  & 0  & 0  & 0  & 0  & 0  & 0   & 0 \\ \hline
			2 & 11  & 1688 & 194  & 105 & 0  & 0  & 0  & 0  & 0  & 0   & 0  \\ \hline
			3 & 0   & 153  &1653 &157  &37   & 0  & 0  & 0  & 0  & 0   & 0 \\ \hline
			4 & 0   & 9   & 346 &1580 & 65 & 0  & 0  & 0  & 0  & 0   & 0 \\ \hline
			5 & 0   & 9   & 31  &102  &1649 &209  & 0  & 0  & 0  & 0   & 0 \\ \hline
			6 & 0   & 0   & 0  &11   & 105 &1651 & 233 & 0  & 0  & 0   & 0 \\ \hline
			7 & 0   & 0   & 0  & 0  & 8  &159  &1694 & 139  & 0  & 0   & 0 \\ \hline
			8 & 0   & 0   & 0  & 0  & 0  & 0  & 64  &1910 & 26  & 0   & 0 \\ \hline
			9 & 0   & 0   & 0  & 0  & 0  & 0  & 0  &35  &1798 & 167   & 0 \\ \hline
			10& 0   & 0   & 0  & 0  & 0  & 0  & 0  & 21  &105  &1853  & 21 \\ \hline
			11& 0   & 0   & 8  & 4  & 0  & 4  & 0  & 1  & 0  &143    & 1840 \\ \hline
		\end{tabular}
		\label{confmatrixjpeg}
	}
	\subtable[Down-sampling level class]{
		\begin{tabular}{|p{0.3cm}<{\centering}|p{0.3cm}<{\centering}|p{0.3cm}<{\centering}|p{0.3cm}<{\centering}|p{0.3cm}<{\centering}|p{0.3cm}<{\centering}|p{0.3cm}<{\centering}|p{0.3cm}<{\centering}|p{0.3cm}<{\centering}|p{0.3cm}<{\centering}|p{0.3cm}<{\centering}|p{0.3cm}<{\centering}|}
			\hline
			\diagbox{}{} & 1   & 2   &3   & 4  & 5  & 6  & 7  & 8  & 9  & 10  & 11 \\ \hline
			1 & 2000 & 0  & 0  & 0  & 0  & 0  & 0  & 0  & 0  & 0   & 0 \\ \hline
			2 & 9   & 1901 & 64  & 20  & 6  & 0  & 0  & 0  & 0  & 0   & 0  \\ \hline
			3 & 3   & 69   &1891 &31   &6   & 0  & 0  & 0  & 0  & 0   & 0 \\ \hline
			4 & 0   & 0   & 64  &1853 & 76  & 5  & 0  & 0  & 0  & 0   & 0 \\ \hline
			5 & 0   & 1   & 0  & 41  &1899 &58   & 1  & 0  & 0  & 0   & 0 \\ \hline
			6 & 0   & 0   & 0  &5   & 61  &1907 & 21  & 6  & 0  & 0   & 0 \\ \hline
			7 & 0   & 0   & 0  & 0  & 0  & 62  &1913 & 25  & 0  & 0   & 0 \\ \hline
			8 & 0   & 0   & 0  & 0  & 0  & 26  & 54  &1867 & 53  & 0   & 0 \\ \hline
			9 & 0   & 0   & 0  & 0  & 0  & 0  & 0  &59  &1895  & 46   & 0 \\ \hline
			10& 0   & 0   & 0  & 0  & 0  & 0  & 0  & 0  &61  &1897  & 42 \\ \hline
			11& 0   & 0   & 0  & 0  & 0  & 0  & 0  & 0  & 37  &42    & 1921 \\ \hline
		\end{tabular}
		\label{confmatrixresolution}
	}
	\caption{Confusion matrix of quality level estimation using the proposed model (Five Conv. layers).}
	\label{confmatrixquality}
\end{table*}

\subsection{Quality classified face detection and recognition}
\subsubsection{\textbf{Evaluation protocol and datasets}}
We use face detection and recognition as specific applications to evaluate our proposed image quality classified image detection and recognition framework. We first test different methods on each low-quality dataset to compare their performance and then test our framework on a mix-quality dataset, which consists of images that are randomly decreased into one of the quality classes.  


We used AFLW \cite{koestinger2011annotated} as the face detection dataset, which contains 25,993 faces in 21,997 images. The face recognition is tested on the CASIA-Webface \cite{Yi2014Learning} dataset, which consists of 494,414 faces from 10,575 subjects. The datasets are separated into training and testing sets with a ratio of 0.8:0.2. In the first experiment, all images are processed into all the quality classes we previously defined. In the second experiment, the \textbf{mix-quality} set is prepared by randomly decreasing each image into one of the quality classes. It is then separated into a mix-quality training set and a mix-quality testing set.

\subsubsection{\textbf{Performance on low quality sets}}

\begin{figure*}{}
	\centering
	\includegraphics[angle=0,scale=0.28]{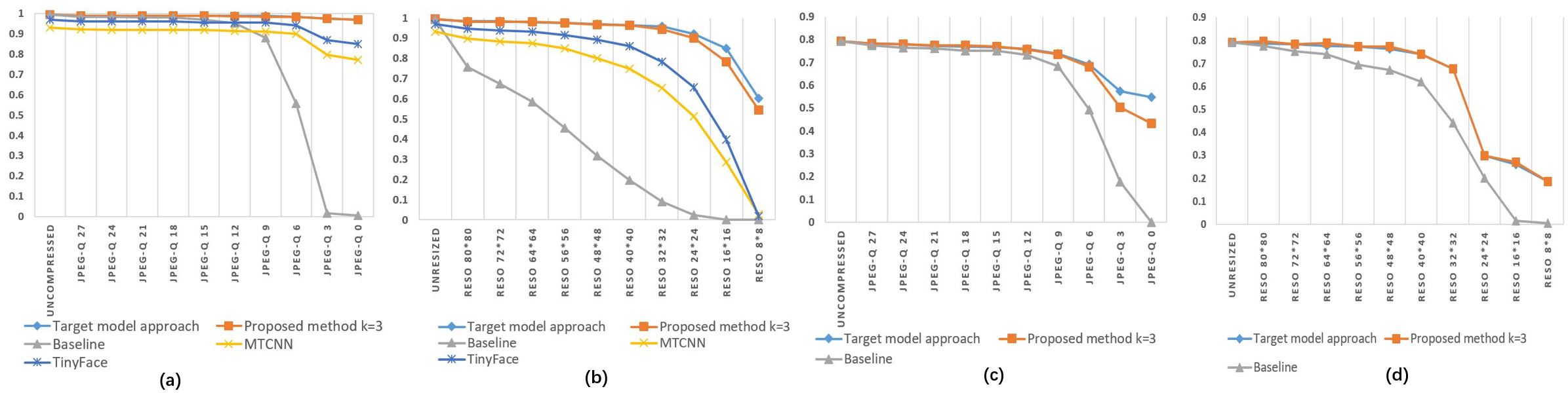}
	\centering
	\caption{Performance reduction on low-quality images. (a) Face detection results with JPEG compression distortion. (b) Face detection results with down-sampling images. (c)Face recognition results with JPEG compression distortion.  (d) Face recognition results with down-sampling images.}
	\label{performancereduction}
\end{figure*}

In this experiment, we exam how will image quality affect face detection and recognition. 
In the face detection application, we adopted three settings. \textbf{Setting 1:} train on high quality, test on low-quality level images. We define the \textbf{baseline} as Faster RCNN \cite{Sun2017Face} implementation for face detection, we also applied two other popular face detection tools, MTCNN \cite{zhang2016joint} and TinyFace \cite{hu2016finding}. For all the three methods, we train these face detectors on the unprocessed original data and test them on each level of the processed data, either JPEG compression or down-sampling. \textbf{Setting 2:} train and test on the same quality class dataset. We define \textbf{target models approach} as the Faster RCNN approach that train and test on each quality class dataset separately, i.e., train 11 target detectors and test them on the corresponding quality class testing data. \textbf{Setting 3 (proposed framework):} predict the quality type and severity class, fuse detection results coming from corresponding detectors. We follow the proposed framework and test it on each of the 11 classes testing data. Again, the Faster RCNN approach is chosen to train the 11 models seperately. After quality prediction, $K=3$ models are chosen for fusion. The results are denoted as \textbf{proposed method} and plotted in the Figure.\ref{performancereduction} (a) and (b).


From the results, it is seen that the performance of all three methods, baseline, MTCNN and TinyFace in setting 1, drop dramatically when the corresponding JPEG compression and down-sampling levels reach to a certain level. In setting 2, if the quality class is the given prior knowledge, i.e., the method can select the correct model to analyze the image, the results can be improved significantly. Again, it proved that it is not the image quality itself that is the most important, but rather the quality of the images used to train the analyzer should be similar to that of those on which the analyzer will be tested. In the last setting, we can further prove, even without the quality information, our proposed framework can estimate the quality well and fuse the right target models to achieve a promising result.

For the face recognition application, VGG\_Face \cite{parkhi2015deep} is chosen as the baseline evaluation method. We follow the same three settings as previously, and denote them as baseline, target model approach and proposed method, respectively. As shown in Figure. \ref{performancereduction}(c) and (d), the face recognition application obtain similar results.

\subsubsection{\textbf{Results on the mix-quality dataset}}

\begin{table}
	\centering
	\footnotesize
	\newcommand{\tabincell}[2]{\begin{tabular}{@{}#1@{}}#2\end{tabular}}
	\begin{tabular}{|c|c|c|}
		\hline
		\tabincell{c}{Training\\ setting} & \tabincell{c}{Face detection\\ methods}  & Accuracy(mAP) \\ \hline
		\multirow{3}{*}{\tabincell{c}{Standard\\ (quality unknown)}}
		& MTCNN                  & 0.7541      \\ \cline{2-3}
		& TinyFace               & 0.7310    \\ \cline{2-3}
		&\tabincell{c}{Faster RCNN}  & 0.7292     \\ \hline
		\tabincell{c}{Mixed-quality \\ (quality unknown)} 	& \tabincell{c}{Faster RCNN} & 0.9095 \\ \hline
		
		\tabincell{c}{Target model \\  (qulity known)} 	& \tabincell{c}{Faster RCNN} & 0.9557 \\ \hline

		\multirow{3}{*}{\tabincell{c}{Target model\\(quality predicted)}}
		& \tabincell{c}{Our method* (K=1)}     & 0.9216      \\ \cline{2-3}
		&\tabincell{c}{Our method* (K=3)}        & 0.9512      \\ \cline{2-3}
		&\tabincell{c}{Our method* (K=5)}        & 0.9602     \\ \hline
	\end{tabular}
	\caption{	\footnotesize Face detection accuracy on mix-quality dataset. Our method * denotes Faster RCNN is applied within our proposed framework.}\label{frameworktest}
\end{table}

\begin{table}
	\centering
	\footnotesize
	\newcommand{\tabincell}[2]{\begin{tabular}{@{}#1@{}}#2\end{tabular}}
	\begin{tabular}{|c|c|c|}
		\hline
		\tabincell{c}{Training\\setting}&\tabincell{c}{Face recognition\\ methods}  & Accuracy \\ \hline
		\tabincell{c}{Standard\\(quality unknown)} 
		&\tabincell{c}{VGG\_face}  & 61.4\%    \\ \hline
		\tabincell{c}{Mixed-quality\\(quality unknown)} & \tabincell{c}{VGG\_face} & 63.43\% \\ \hline
		\tabincell{c}{Target model \\ (quality known)} 	& \tabincell{c}{VGG\_face} & 65.65 \\ \hline
		\multirow{3}{*}{\tabincell{c}{Target model\\(quality predicted)}}
	   &  \tabincell{c}{Our method\# (K=1)}   & 62.63\%      \\ \cline{2-3}
		& \tabincell{c}{Our method\# (K=3)}    & 65.02\%      \\ \cline{2-3}
		& \tabincell{c}{Our method\# (K=5)}    & 65.81\%     \\ \hline
	\end{tabular}
	\caption{\footnotesize Face recognition accuracy on mix-quality dataset. Our method\# denotes VGG\_face is applied within our proposed framework.}\label{frameworktest2}
\end{table}
We intend to simulate how the face detection/recognition techniques perform in a real-world scenario by applying different training settings and test on the mix-quality dataset. These training settings involve the standard dataset (high-quality images) training, the mix-quality dataset training, the target model training, and our proposed framework fusing results coming from $K$ predicted target models. 


As shown in TABLE \ref{frameworktest}, we train the models in different settings and test them on the mixed-quality testing dataset. We apply Faster RCNN as the baseline face detection method in our proposed image quality classified framework, which predict the quality of the image and fuse the results coming from target models trained on each class level separately. As a comparison, we also present results of MTCNN, TinyFace and Faster RCNN in different settings. In the real world scenario, the result shows, if we train the models on a good quality dataset, we get a low accuracy in all three state-of-the-art face detection methods, namely, MTCNN, TinyFace and Faster RCNN. If we switch the training dataset with a mixed-quality one, we could improve the accuracy from around 0.73 to 0.9. Ideally, if we know the quality of each testing image precisely and selecting target model training on the corresponding quality level data, we can increase the result to 0.955. It is not realistic to know the quality class of each image beforehand in the real world scenario. However, our proposed method could help to predict the quality classes as well as their severity, together with model fusion, we can further improve the result to 0.96 if we choose top 5 models to fuse. TABLE \ref{frameworktest2} shows our proposed framework achieves similar performance improvements in face recognition.  

Hence, through extensive experimental results, we show that image quality factor has a great impact on the face detection/recognition applications. Through carefully designed quality prediction network, we could recognize poor quality image caused by either JPEG compression or low resolution with high accuracy and efficiency. Using the proposed model fusion framework, we can significantly boost the accuracy of face detection/recognition in the real-world scenario.

\section{Conclusion}
In this work, we have highlighted the image quality issue in image-based object recognition and detection. We have presented an image quality classifed image analysis framework to reduce the effect of image quality factor on the performances of object detection and recognition systems. We have shown that deep learning neural networks can recognize the type and severity image quality degradations. We show that by designing face detecting and recognition systems for the correct quality image classes, we can significantly improve image analysis system's performances.  

\section{Acknowledgement}
The author acknowledges the financial support from the International Doctoral Innovation Centre, Ningbo Education Bureau, Ningbo Science and Technology Bureau, and the University of Nottingham. This work was also supported by the UK Engineering and Physical Science Research Council [grant number EP/L015463/1].

\bibliographystyle{plain}
\bibliography{reference}






%

\end{document}